\documentclass[a4paper,fleqn, review]{cas-sc}

\usepackage[numbers]{natbib}
\usepackage[utf8]{inputenc}
\usepackage[T1]{fontenc}
\def\tsc#1{\csdef{#1}{\textsc{\lowercase{#1}}\xspace}}
\tsc{WGM}
\tsc{QE}
\tsc{EP}
\tsc{PMS}
\tsc{BEC}
\tsc{DE}
\usepackage{epigraph}
\usepackage{longtable}
\usepackage{setspace}

\usepackage{textcomp}

\usepackage{multirow}

\usepackage{color}

\usepackage{svg}
\usepackage{amsmath}
\definecolor{WordGreen}{RGB}{100, 136, 40}
\definecolor{WordDarkGrey}{RGB}{82, 82, 82}
\definecolor{WordRed}{RGB}{192, 80, 77}
\definecolor{WordBlue}{RGB}{0, 122, 192}

\definecolor{WordLightBlue}{RGB}{218, 238, 243}
\definecolor{WordLightGreen}{RGB}{234, 241, 221}

\definecolor{WordFillGreen}{RGB}{194, 214, 155}
\definecolor{WordFillRed}{RGB}{252, 214, 182}
\definecolor{WordFillGray}{RGB}{217, 217, 217}

\UseRawInputEncoding
\usepackage{times}
\usepackage{epsfig}
\usepackage{amssymb}
\usepackage[ruled,vlined,linesnumbered]{algorithm2e}
\usepackage{comment}
\usepackage{array, makecell} 
\usepackage{lettrine}
\usepackage{url} 
\usepackage{lscape}
\usepackage{cuted}
\usepackage{tabularx}
\usepackage{colortbl}
\usepackage[normalem]{ulem}
\usepackage{hhline}
\usepackage{vcell}
\usepackage{longtable}
\usepackage{vcell}
\usepackage{rotating}
\usepackage{pdflscape}
\usepackage{sidecap}
\usepackage{neuralnetwork}
\usepackage[export]{adjustbox} % https://tex.stackexchange.com/questions/91566/syntax-similar-to-centering-for-right-and-left
\usepackage[accsupp]{axessibility} % Improves PDF readability for those with disabilities.
\usepackage{amsfonts,bm}
\usepackage{physics}
\usepackage{wrapfig}
\usepackage{caption}
\usepackage{color,soul}
\usepackage[frozencache,cachedir=minted-cache]{minted}
\usepackage{mathtools}
\usepackage{amsthm}

\usepackage{acronym}
\acrodef{FCN}[FCN]{Fully Convolutional Network}
\acrodef{GAME}[GAME]{Grid Average Mean Absolute Error}
\acrodef{DL}[DL]{Deep Learning}
\acrodef{DNN}[DNN]{Deep Neural Network}
\acrodef{ML}[ML]{Machine Learning}
\acrodef{CV}[CV]{Computer Vision}
\acrodef{AI}[AI]{Artificial Intelligence}
\acrodef{CNN}[CNN]{Convolutional Neural Network}
\acrodef{RNN}[RNN]{Recurrent Neural Network}
\acrodef{GAN}[GAN]{Generative Adversarial Network}
\acrodef{JCU}[JCU]{James Cook University}
\acrodef{MAE}[MAE]{Mean Average Error}
\acrodef{MAP}[mAP]{Mean Average Precision}
\acrodef{CA}[CA]{Classification Accuracy}
\acrodef{LCFCN}[LCFCN]{Localization-based Counting loss Fully Convolutional Network}
\acrodef{IoT}[IoT]{Internet of Things}
\acrodef{MLP}[MLP]{Multi-Layer Perceptrons}
\def\x{{\mathbf x}}				% Definition of the bold letter x
\def\y{{\mathbf y}}				% Definition of the bold letter y
\def\vtheta{{\bm{\theta}}}
\usepackage{xspace}

\DeclareMathOperator*{\argmin}{arg\,min}

\usepackage[capitalize]{cleveref}
\crefname{section}{Sec.}{Secs.}
\Crefname{section}{Section}{Sections}
\Crefname{table}{Table}{Tables}
\crefname{table}{Table}{Table}
\usepackage{subcaption}
\usepackage{booktabs}
\usepackage{arydshln}
\usepackage{listings}
\definecolor{codegreen}{rgb}{0,0.6,0}
\definecolor{codegray}{rgb}{0.5,0.5,0.5}
\definecolor{codepurple}{rgb}{0.58,0,0.82}
\definecolor{backcolour}{rgb}{0.95,0.95,0.92}
\definecolor{cleacolorr}{rgb}{1,1,1}

\lstdefinestyle{mystyle}{
    backgroundcolor=\color{cleacolorr},   
    commentstyle=\color{codegreen},
    keywordstyle=\color{magenta},
    numberstyle=\tiny\color{codegray},
    stringstyle=\color{codepurple},
    basicstyle=\ttfamily\footnotesize,
    breakatwhitespace=false,         
    breaklines=true,                 
    captionpos=b,                    
    keepspaces=true,                 
    numbers=left,                    
    numbersep=5pt,                  
    showspaces=false,                
    showstringspaces=false,
    showtabs=false,                  
    tabsize=2
}

\usepackage[most]{tcolorbox}
\newtcolorbox[auto counter]{pabox}[2][]{%
colback=blue!5!white,colframe=blue!75!black,fonttitle=\bfseries,
title=Box~\thetcbcounter: #2,#1}
\usepackage{tikz}
\usepackage{graphicx}
\graphicspath{{./Graphics/}}
\DeclareGraphicsExtensions{.pdf,.jpeg,.png,.jpg}

\usepackage{amsmath}

\usepackage{array}

\usepackage{url}
\begin{document}
% \doublespacing
% \linenumbers

\let\WriteBookmarks\relax
\def\floatpagepagefraction{1}
\def\textpagefraction{.001}

\shorttitle{Weed Detection in Challenging Field Conditions}

\shortauthors{Saleh et~al.}

\title[mode = title]{Weed Detection in Challenging Field Conditions: A~Semi-Supervised Framework for Overcoming Shadow Bias and Data Scarcity}

\author[1]{Alzayat Saleh}[orcid=0000-0001-6973-019X]
\cormark[1]
\ead{alzayat.saleh@my.jcu.edu.au}
% % Email id of the first author
% \ead{alzayat.saleh@my.jcu.edu.au}
\credit{Conceptualisation, Data Curation, Data Analysis, Supervision, Reviewing/editing the draft}

\author[1]{Shunsuke Hatano}%[orcid=0000-0001-6973-019X]

% \ead{shunsuke.hatano@jcu.edu.au}
\credit{Conceptualisation, Data Curation, Data Analysis, Software Development, DL Algorithm Design, Visualization, Writing original draft}

\author[1,2,3]{Mostafa Rahimi Azghadi}[                        orcid=0000-0001-7975-3985]

\credit{Conceptualisation, Data Curation, Data Analysis, Supervision, Reviewing/editing the draft}

\affiliation[1]{organization={College of Science and Engineering, James Cook University},
    % addressline={Radarweg 29}, 
    city={Townsville},
    % citysep={}, % Uncomment if no comma needed between city and postcode
    postcode={4814}, 
    state={QLD},
    country={Australia}}

\affiliation[2]{organization={Agriculture Technology and Adoption Centre, James Cook University},
    % addressline={Radarweg 29}, 
    city={Townsville},
    % citysep={}, % Uncomment if no comma needed between city and postcode
    postcode={4814}, 
    state={QLD},
    country={Australia}}

    \affiliation[3]{organization={ARC Training Centre in Plant Biosecurity},
    % addressline={Radarweg 29}, 
    %city={Townsville},
    % citysep={}, % Uncomment if no comma needed between city and postcode
    %postcode={4814}, 
    %state={QLD},
    country={Australia}}

\cortext[cor1]{Corresponding author}
% \cortext[cor2]{Principal corresponding author}

\begin{abstract}
The automated management of invasive weeds is critical for sustainable agriculture, yet the performance of deep learning models in real-world fields is often compromised by two factors: challenging environmental conditions and the high cost of data annotation. This study tackles both issues through a diagnostic-driven, semi-supervised framework. Using a unique dataset of approximately 975 labeled and 10,000 unlabeled images of Guinea Grass in sugarcane, we first establish strong supervised baselines for classification (ResNet) and detection (YOLO, RF-DETR), achieving F1 scores up to 0.90 and mAP50 scores exceeding 0.82.
Crucially, this foundational analysis, aided by interpretability tools, uncovered a pervasive "shadow bias," where models learned to misidentify shadows as vegetation. This diagnostic insight motivated our primary contribution: a semi-supervised pipeline that leverages unlabeled data to enhance model robustness. By training models on a more diverse set of visual information through pseudo-labeling, this framework not only helps mitigate the shadow bias but also provides a tangible boost in recall, a critical metric for minimizing weed escapes in automated spraying systems. To validate our methodology, we demonstrate its effectiveness in a low-data regime on a public crop-weed benchmark. Our work provides a clear and field-tested framework for developing, diagnosing, and improving robust computer vision systems for the complex realities of precision agriculture.
\end{abstract}

% Keywords
\begin{keywords}
Semi-supervised Learning, \sep
Weed Detection, \sep
Deep Learning, \sep
% Multi-scale Representation, \sep
Computer Vision, \sep
Agriculture. \sep
% polariton \sep \WGM \sep \BEC
\end{keywords}

\maketitle

% \tableofcontents
%%%%%%%%%%%%%%%%%%%%%%%%%%%%%%%%%%%%%%%%%%%%%%%%%%%%%%%%%%%%%%%%

%%%%%%%%%
%%%%%%%%%%%%%%%%%%%%%%%%%%%%%%%%%%%%%%%%%%%%%%%%%%%%%%%%%%%%%%%%
\section{Introduction}\label{sec:intro}

Weeds pose a persistent threat to global crop production, requiring extensive resources for manual removal or chemical intervention~\cite{oerke2006}. In the pursuit of more sustainable and efficient agricultural practices, precision agriculture has emerged as a transformative paradigm. By leveraging advanced sensing and automation, it enables targeted interventions that can significantly reduce the economic and environmental costs associated with weed management. Central to this approach is the development of robust, real-time weed detection systems capable of operating effectively in diverse and dynamic field conditions \cite{azghadi2025precision, lammie2019low}.

The emergence of deep learning has revolutionized computer vision, and its application to weed detection is no exception. Convolutional Neural Networks (CNNs), particularly single-stage object detectors like \emph{You Only Look Once} (YOLO)~\cite{redmon2018}, have been widely adopted for their ability to balance high accuracy with real-time inference speeds, making them ideal for deployment on agricultural robots. Concurrently, newer architectures based on the Transformer model, such as the Detection Transformer (DETR)~\cite{carion2020} and its real-time variants~\cite{guo2024,yang2023}, offer a compelling end-to-end alternative that removes the need for many hand-designed components \cite{saleh2024weedclr}.

Despite these advancements, the practical deployment of such models in real-world agricultural environments reveals critical challenges that are often abstracted away in benchmark datasets. First, model performance can degrade substantially under the variable lighting, shadows, and occlusions inherent to field operations \cite{saleh2025fieldnet}. Our preliminary work identified that models can develop a strong "shadow bias," learning to associate high-contrast shadows with vegetation rather than relying on true morphological features. Second, the supervised learning paradigm, upon which these models are built, demands massive, meticulously annotated datasets. The manual labeling process is not only a significant financial and temporal bottleneck but is also prone to error, with studies reporting that even trained agronomists can exhibit mislabeling rates as high as 12\%~\cite{dyrmann2016}.

\emph{Semi-supervised learning} (SSL) has gained traction as a powerful strategy to mitigate this dependency on labeled data by leveraging the vast quantities of unlabeled imagery that are easy to collect~\cite{vanengelen2020}. While SSL has shown promise in agricultural segmentation tasks~\cite{perezortiz2015,nong2022}, and techniques like pseudo-labeling have been shown to boost recall in weed detection~\cite{saleh2025semi}, a systematic comparison of how SSL impacts different object detection architectures (CNN vs. Transformer) in challenging field conditions remains an open question.

This paper addresses these gaps by presenting a comprehensive framework for weed detection that tackles the dual problems of environmental variability and data scarcity. We report a systematic investigation that begins with supervised learning, uses interpretability to diagnose critical failure modes, and leverages semi-supervised learning as a practical solution. Our primary contributions are as follows:
\begin{itemize}
    \item We propose a {dual-pipeline comparative framework} that systematically benchmarks both quadrant-based classification (ResNet) and object detection (YOLO, RF-DETR) to evaluate their respective strengths in a real-world scenario.
    \item We identify and diagnose {"shadow bias"} as a critical, yet often overlooked, failure mode in agricultural vision systems. We use interpretability tools (Grad-CAM) to visualize this phenomenon, guiding the development of more robust models.
    \item We implement and evaluate a {practical semi-supervised learning strategy} that integrates unlabeled data through single-pass pseudo-labeling, demonstrating its effectiveness in improving recall and generalization under challenging field conditions.
    \item We provide a {rigorous and transparent methodology}, including the explicit handling of data leakage issues discovered during experimentation, offering a robust comparison of state-of-the-art architectures on a unique, field-collected dataset.
\end{itemize}

The remainder of this paper is structured as follows: Section~\ref{rltdwork} reviews existing literature on weed detection and semi-supervised learning. Section~\ref{experiments} details our experimental setup and datasets. Section~\ref{results} presents our empirical findings, followed by an in-depth discussion in Section~\ref{sec:discussion}. Finally, Section~\ref{conclusions} summarizes our key contributions and their implications for precision agriculture.
%%%%%%%%%%%%%%%%%%%%%%%%%%%%%%%%%%%%%%%%%%%%%%%%%%%%%%%%%%%%%%%%

%%%%%%%%%%%%%%%%%%%%%%%%%%%%%%%%%%%%%%%%%%%%%%%%%%%%%%%%%%%%%%%%
\section{Related Work} \label{rltdwork}

The automated identification of weeds has been a longstanding goal in precision agriculture, with methodologies evolving in sophistication alongside advances in computer vision and deep learning. This section reviews the key technological paradigms, from classical techniques to modern deep learning architectures, and situates our work within the current research landscape.

\subsection{Classical and Early Machine Learning Approaches}
Early attempts at automated weed detection relied on classical computer vision techniques, which typically involved a two-step process of feature engineering followed by classification. These methods often utilized hand-crafted features based on color, shape, and texture. For instance, color-space transformations and thresholding of vegetation indices were common for segmenting plants from soil backgrounds~\cite{Torres2015}. Subsequent classification was performed using traditional machine learning algorithms like Support Vector Machines (SVMs) or Random Forests~\cite{Wang2019}. While these methods were computationally efficient and required relatively small datasets, their performance was fragile. They often failed to generalize across varying field conditions, being highly sensitive to changes in illumination, soil type, and plant growth stages~\cite{Hasan2021,Wang2019}. Their reliance on manual feature engineering made them fragile and unable to capture the complex hierarchical patterns needed for robust weed identification.

\subsection{Supervised Deep Learning for Weed Detection}
The deep learning revolution marked a paradigm shift in weed detection, with Convolutional Neural Networks (CNNs) becoming the dominant approach. CNNs eliminate the need for manual feature engineering by learning discriminative representations directly from pixel data. Initial work focused on image-level classification, where models like Inception and ResNet were trained to distinguish between different weed species and crops~\cite{Ferreira2017}. The creation of large-scale public datasets, such as {DeepWeeds}~\cite{Olsen2019}, was instrumental in this progress, enabling models to achieve classification accuracies exceeding 95\% on multi-species tasks. Architectures like {ResNet}~\cite{He2016}, in particular, became a staple in agricultural vision due to their ability to train very deep networks effectively.

More recently, the focus has shifted from classification to the more challenging task of {object detection}, which provides the precise location of each weed instance required for targeted spraying. Two-stage detectors like Faster R-CNN have been successfully applied but often at the expense of real-time performance~\cite{saleem2022weed}. Consequently, single-stage detectors like \emph{You Only Look Once} (YOLO) and its variants have gained prominence for their impressive balance of speed and accuracy~\cite{Dang2023}. Other single-stage models like {RetinaNet} have also been explored, particularly for their use of a focal loss to handle the class imbalance between sparse weeds and dense crops~\cite{peng2022weed}. While these CNN-based detectors have set high benchmarks, they can still struggle with detecting very small or heavily occluded weeds.

\subsection{Transformer-Based Architectures in Agriculture}
Inspired by their success in natural language processing, Transformer-based architectures have recently been adapted for computer vision. The Detection Transformer (DETR)~\cite{carion2020} introduced a fully end-to-end object detection pipeline, eliminating the need for hand-tuned components like anchors and non-maximum suppression (NMS). While pioneering, the original DETR suffered from slow convergence and high computational cost. Subsequent research has focused on developing more efficient variants. In agriculture, DETR-like models are beginning to be explored for tasks like weed detection~\cite{Liao2025,Islam2025}. For instance, {RF-DETR}~\cite{Roboflow2025} represents a recent effort to create a real-time, high-performance Transformer-based detector. However, the application and systematic comparison of these models in high-similarity crop-weed environments remain emergent, with few studies directly benchmarking their performance against highly optimized CNNs under identical conditions.

\subsection{Semi-Supervised Learning to Reduce Annotation Burden}
A significant and practical barrier to deploying deep learning models in agriculture is the substantial cost and effort required for data annotation~\cite{Liu2023}. Semi-supervised learning (SSL) directly addresses this by leveraging large amounts of unlabeled data alongside a smaller labeled set. A common and effective SSL technique is {pseudo-labeling}, where a "teacher" model trained on labeled data generates annotations for unlabeled images. These pseudo-labeled samples are then used to train a "student" model, enhancing its generalization. In agriculture, SSL has been successfully used to improve model accuracy while drastically reducing labeling requirements~\cite{saleh2025semi,Shorewala2021}. However, much of the existing work has focused on classification or segmentation tasks. The systematic application of SSL to different object detection architectures (i.e., YOLO vs. DETR) for weed detection, and an analysis of its impact on robustness against real-world challenges like "shadow bias," is an underexplored area.

Our work is positioned at the intersection of these research directions. We provide a direct, rigorous comparison of state-of-the-art CNN (YOLO) and Transformer (RF-DETR) architectures on a challenging, field-collected dataset. Crucially, we extend this comparison to the semi-supervised domain, evaluating how pseudo-labeling impacts the performance and robustness of each architectural paradigm, thereby addressing a key gap in the current literature.
%%%%%%%%%%%%%%%%%%%%%%%%%%%%%%%%%%%%%%%%%%%%%%%%%%%%%%%%%%%%%%%%

%%%%%%%%%%%%%%%%%%%%%%%%%%%%%%%%%%%%%%%%%%%%%%%%%%%%%%%%%%%%%%%%%%%%%%%%%%%%%%%%%
% ######################################################

\begin{figure}[ht!]
    \centering
    \includegraphics[width=0.98\textwidth]{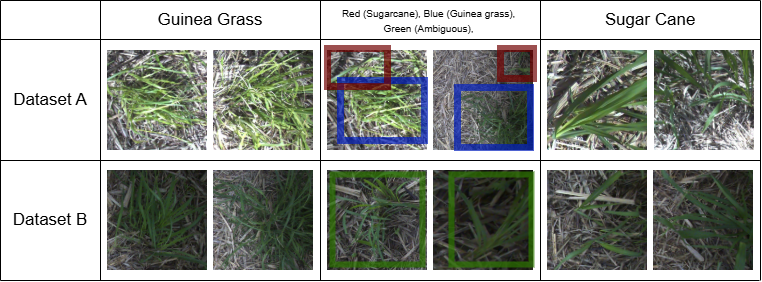}
    \caption{Comparison of Dataset~A (top row) vs.\ Dataset~B (bottom row).
    Shadows and color overlap make weed detection more difficult in B.
    Each column shows examples of (1) Guinea Grass, (2) Both/Ambiguous, and (3) Sugar Cane.}
    \label{fig:visdiff}
\end{figure}

\section{Experiments}\label{experiments}

Our experimental methodology was designed to systematically evaluate and compare different approaches for weed detection under realistic and challenging field conditions. We structured our investigation into two primary pipelines: (1) a quadrant-based binary {classification} approach to assess broad area presence, and (2) a bounding-box-based {object detection} approach for precise localization. % Both pipelines were benchmarked under fully supervised and semi-supervised paradigms, as illustrated in Figures~\ref{fig:pipeline-class} and~\ref{fig:pipeline-detect}.

\subsection{Dataset Curation and Preparation} \label{sec:datasets}

A key contribution of our work is the use of a carefully curated dataset that reflects the complexities of real-world agriculture. All data was collected from sugarcane paddocks heavily infested with Guinea Grass.

\paragraph{Labeled Datasets: A and B.}
Our initial analysis revealed that model performance was highly dependent on environmental conditions. To study this systematically, we partitioned our labeled data into two distinct sets, as shown in Figure~\ref{fig:visdiff}:
\begin{itemize}
    \item {Dataset A:} Comprising images from paddocks `mw5\_1330` and `mw5\_1331`, this set features relatively clear, well-lit conditions, serving as our baseline for model performance.
    \item {Dataset B:} A more challenging compilation including images from `mw5\_1327` and `paddock\_wt2`. This set is characterized by darker images, strong, inconsistent shadows, and higher visual similarity between crop and weed, designed specifically to test model robustness under adverse conditions.
\end{itemize}
Together, these sets contain approximately {975 labeled images}. The detailed distribution of annotations is presented in Table~\ref{tab:boxcounts}.

\begin{table}[ht]
\centering
\caption{Bounding-box distribution across labeled paddocks in Datasets~A and B.}
\label{tab:boxcounts}
\begin{tabular}{lrr}
\toprule
\textbf{Paddock ID} & \textbf{Sugarcane} & \textbf{Guinea Grass} \\
\midrule
\texttt{paddock\_A1} & 3605 & 239 \\
\texttt{paddock\_A2} & 837  & 112 \\
\midrule
\textbf{Dataset A}   & 4442 & 351 \\
\midrule
\texttt{paddock\_B1} & 170  & 29  \\
\texttt{paddock\_B2} & 336  & 252 \\
\midrule
\textbf{Dataset B}   & 506  & 281 \\
\bottomrule
\end{tabular}
\end{table}

\begin{figure}[ht]
    \centering
    \includegraphics[width=0.98\textwidth]{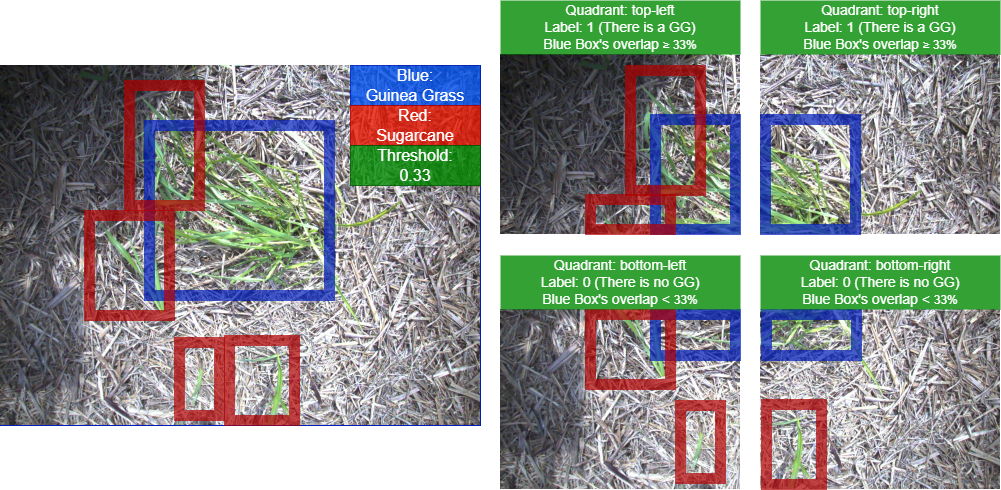}
    \caption{Quadrant splitting and labeling with a 33\% overlap threshold.
    The original image (left) has bounding boxes for Guinea Grass (blue) and Sugarcane (red).
    Each quadrant (right) is saved as a separate image. A quadrant is labeled `1' (weed-positive)` if at least one GG box overlaps $\geq 33\%$ of its bounding box area; otherwise `0' (weed-negative)`.}
    \label{fig:quad-explain}
\end{figure}

\paragraph{Quadrant Splitting and Label Generation for Classification.}
For the classification task, each high-resolution source image \(\mathbf{I}\) was divided into four non-overlapping quadrants, \(\{\text{quadrant}_i\}_{i=1}^4\). A quadrant was assigned a binary label \(y_i \in \{0, 1\}\) based on the presence of Guinea Grass (\textit{GG}). A quadrant was labeled as positive (\(y_i=1\)) if any single \textit{GG} bounding box within it satisfied the following condition:
\begin{equation}
  \frac{\text{Area}(\text{box}_{gg} \cap \text{quadrant}_i)}{\text{Area}(\text{box}_{gg})} \geq \tau
\end{equation}
where the overlap threshold \(\tau\) was set to 0.33. This strategy, visualized in Figure~\ref{fig:quad-explain}, ensures that only quadrants with significant weed presence are considered positive, minimizing noise from marginal overlaps. For the detection task, this quadrant-based approach was found to be detrimental, as splitting often truncated objects and degraded performance. Therefore, all detection models were trained on full, unsplit images.

\paragraph{Handling Class Imbalance.}
As shown in Table~\ref{tab:boxcounts}, the raw dataset exhibited a significant class imbalance, with sugarcane instances far outnumbering Guinea Grass instances. To prevent our models from developing a bias towards the majority class, we employed a downsampling strategy on the training set, removing a subset of images containing only sugarcane. This rebalancing was crucial for improving the recall of the minority class (Guinea Grass), which is the primary target of our detection system.

\paragraph{Dataset Integrity and Validation Strategy.}
All labeled data was partitioned into training (70\%), validation (20\%), and testing (10\%) sets. We adhered to a strict validation protocol across all experiments to ensure the scientific validity of our results. In multi-stage training pipelines, such as those involving pretraining on one dataset and fine-tuning on another, it is critical to prevent any overlap between the final test set and any data used during any training phase. Initial exploratory experiments were revised to guarantee that our final, reported results are evaluated on a completely held-out test set. This rigorous separation ensures that our metrics are a true and unbiased measure of the models' generalization capabilities.

\paragraph{Public Benchmark Dataset for Method Validation}
To validate the generalizability of our proposed semi-supervised learning methodology, we also conduct experiments on a publicly available dataset, the CropAndWeed Image Dataset ~\cite{Steininger2023WACV}. We explicitly note that the characteristics of this dataset differ from our own; CropAndWeed largely contains images of individual seedlings under more controlled lighting, making it ideal for detection tasks. In contrast, our proprietary dataset is designed to simulate complex detection scenarios with high foliage density and severe lighting variations. Therefore, the purpose of this experiment is not to directly compare raw mAP scores between datasets, but to verify that our semi-supervised pipeline provides a consistent performance improvement over a supervised baseline, even on data with different characteristics.

\begin{figure}[ht!]
    \centering
    \includegraphics[width=0.98\textwidth]{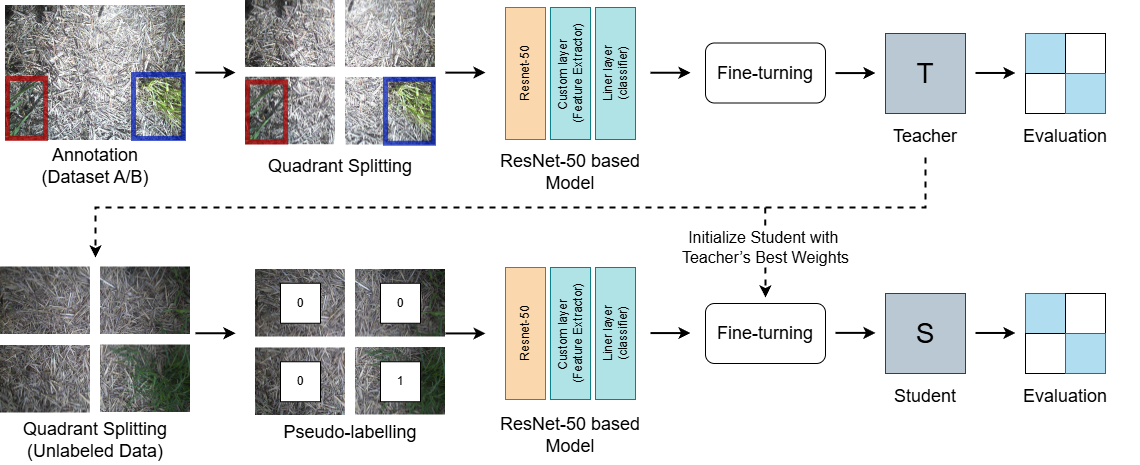}
    \caption{Classification pipeline: each labeled image is split into quadrants, then a ResNet-50 model 
      (initialized from ImageNet) classifies each quadrant as weed or not weed.}
    \label{fig:pipeline-class}
\end{figure}

\subsection{Supervised Learning Pipelines}

\paragraph{Model Architectures.}
Our experiments compare leading architectures from both CNN and Transformer families. For classification, we used a {ResNet-50}~\cite{He2016} backbone. For detection, we benchmarked {YOLOv12-s}, a highly optimized and lightweight CNN-based detector, against {RF-DETR}~\cite{Roboflow2025}, a state-of-the-art real-time Transformer-based detector. The RF-DETR model used was the `RF-DETR-base`, which has a comparable complexity to a medium-sized YOLO model. The overall structure of this supervised pipeline is illustrated in Figure~\ref{fig:pipeline-class}.

\paragraph{Implementation and Hyperparameter Optimization.}
Models were implemented in PyTorch and trained on a distributed system with NVIDIA RTX 3050 and 4060 Ti GPUs. We conducted an extensive hyperparameter search for all models using the {Optuna} framework~\cite{akiba2019optuna}. The search space included learning rate, weight decay, momentum, loss function coefficients, and the intensity of various data augmentations (e.g., flips, rotation degrees, color jitter). Training protocols were tailored to each architecture. The YOLO model was trained for up to {1,000 epochs} with an early stopping mechanism to prevent overfitting. In contrast, the RF-DETR model was trained for approximately {150 epochs} without early stopping, saving the best-performing weights throughout the training process. The general optimization objective for a given model with parameters \(\vtheta\) was to minimize the task-specific loss \(\mathcal{L}\) over the training dataset \(\mathcal{D}_{train}\):
\begin{equation}
  \hat{\vtheta} = \argmin_{\vtheta} \sum_{(\x,\y) \in \mathcal{D}_{train}} \mathcal{L}(f(\x; \vtheta), \y)
\end{equation}
where \(f(\x; \vtheta)\) is the model's prediction for input \(\x\).

\begin{figure}[ht!]
    \centering
    \includegraphics[width=0.98\textwidth]{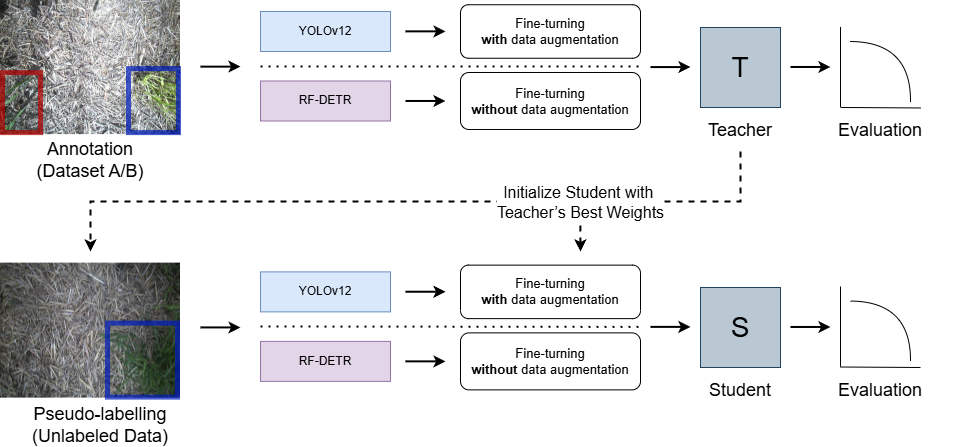}
    \caption{Detection pipeline: YOLOv12-s or RF-DETR is trained on full images with bounding-box annotations.
      Both pipelines eventually finalize with inference and evaluation.}
    \label{fig:pipeline-detect}
\end{figure}

\subsection{Semi-Supervised Learning Pipeline}

To address the annotation bottleneck, we implemented a single-pass pseudo-labeling pipeline as illustrated in Figure~\ref{fig:pipeline-detect}. This approach leverages our \(\sim10,000\) unlabeled images (\(\mathcal{D}_u\)) alongside our smaller labeled set (\(\mathcal{D}_l\)).
\begin{enumerate}
    \item {Train Teacher:} A teacher model, \(f(\x; \hat{\vtheta}_{teacher})\), is first trained exclusively on the labeled set \(\mathcal{D}_l\).
    \item {Generate Pseudo-Labels:} The teacher model is used to predict bounding boxes and class labels, \(\hat{\y}_j\), for each image \(\x_j^u \in \mathcal{D}_u\). Based on our exploratory analysis, predictions were filtered using a class-specific confidence threshold \(c\) (e.g., 0.5 for YOLO, 0.8 for difficult classes in DETR) to ensure high-quality labels. Very small, noisy bounding boxes likely resulting from overfitting were also removed.
    \item {Train Student:} A final student model, \(f(\x; \vtheta_{student})\), is then trained on the combined dataset \(\mathcal{D}_l \cup \mathcal{D}_u\). Its training objective is a weighted combination of the supervised loss on labeled data and the pseudo-supervised loss on unlabeled data:
    \begin{equation}
      \mathcal{L}_{total} = \mathcal{L}_{sup} + \lambda \mathcal{L}_{pseudo} = \frac{1}{N_l} \sum_{i=1}^{N_l} \mathcal{L}_{det}(\x_i^l, \y_i^l) + \lambda \frac{1}{N_u} \sum_{j=1}^{N_u} \mathbb{I}(\hat{p}_j > c) \cdot \mathcal{L}_{det}(\x_j^u, \hat{\y}_j)
    \end{equation}
    where \(\mathcal{L}_{det}\) is the standard detection loss (e.g., from YOLO or DETR), \(\lambda\) is a weighting hyperparameter, \(\mathbb{I}(\cdot)\) is an indicator function that includes an unlabeled sample only if its predicted confidence \(\hat{p}_j\) exceeds the threshold \(c\).
\end{enumerate}

\subsection{Evaluation Metrics}
For the {classification} task, we report accuracy, precision, recall, and the {F1 score}. The F1 score is our primary metric due to the class imbalance present in the data. For the {object detection} task, we use the standard COCO metrics: {mean Average Precision (mAP)} at an IoU threshold of 0.5 (mAP@50) and averaged across IoU thresholds from 0.5 to 0.95 (mAP@50-95). We also closely monitor {precision} and {recall}, as high recall is particularly critical in weed spraying applications to minimize escapes.
%%%%%%%%%%%%%%%%%%%%%%%%%%%%%%%%%%%%%%%%%%%%%%%%%%%%%%%%%%%%%%%%%%%%%%%%%%%%%%%%%
%%%%%%%%%%%%%%%%%%%%%%%%%%%%%%%%%%%%%%%%%%%%%%%%%%%%%%%%%%%%%%%%%%%%%%%%%%%%%%%%%

%%%%%%%%%%%%%%%%%%%%%%%%%%%%%%%%%%%%%%%%%%%%%%%%%%%%%%%%%%%%
\section{Results}\label{results}
In this section, we present the empirical outcomes of our experimental pipelines. We begin by detailing the performance of our supervised classification (SC) models, including a critical interpretability analysis that guided our subsequent work. We then present the comprehensive results from our object detection experiments, comparing fully supervised and semi-supervised approaches across both CNN and Transformer architectures.

\subsection{Fully Supervised Classification}
Our initial investigation focused on a quadrant-based classification pipeline using a ResNet-50 model. We first evaluated the model on the simpler {Dataset A alone} to establish a performance baseline, after which we trained on the more complex {combined A+B dataset} to assess robustness. The final results after extensive hyperparameter tuning are summarized in Table~\ref{tab:class-final}. The model achieved a strong F1 score of 0.88 on the Dataset A test set. When trained on the combined dataset, the model achieved a comparable test F1 of 0.88 from scratch, which was further improved to {0.89} by using the best-performing weights as a pretrained initialization for a final tuning run.

% ----- Classification Results Table -----
\begin{table}[ht]
\centering
\caption{Fully supervised classification  (SC) results on the test set. The model's performance on Dataset A alone reflects its relative simplicity, while fine-tuning on the combined A+B dataset yields the highest overall F1 score.}
\label{tab:class-final}
\begin{tabular}{llccc}
\toprule
\textbf{ID} & \textbf{Dataset(s)} & \textbf{Training Strategy} & \textbf{Val F1} & \textbf{Test F1} \\
\midrule
SC1 & A Only & From Scratch & 0.96 & 0.88 \\
\midrule
SC2 & A+B & From Scratch & 0.86 & 0.88 \\
SC3 & A+B & SC2 as Pretrained Init & 0.86 & \textbf{0.89} \\
\bottomrule
\end{tabular}
\end{table}

\begin{figure}[ht]
    \centering
    \begin{subfigure}[b]{0.48\textwidth}
        \includegraphics[width=\textwidth]{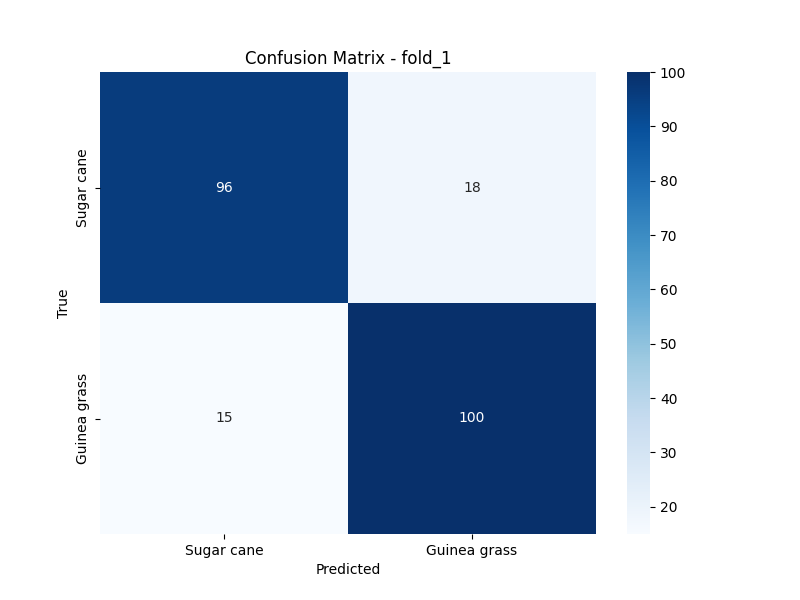}
        \caption{From-scratch training}
    \end{subfigure}
    \hfill
    \begin{subfigure}[b]{0.48\textwidth}
        \includegraphics[width=\textwidth]{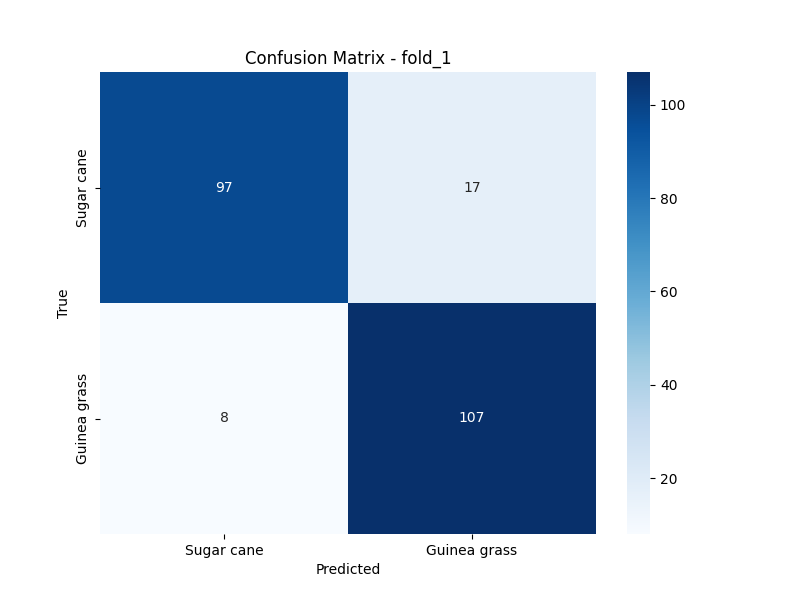}
        \caption{Pretrained initialization}
    \end{subfigure}
    \caption{Confusion matrices on the Dataset A+B test set, comparing (a) from-scratch training with (b) using the best from-scratch model as a pretrained initialization.}
    \label{fig:cmatrix}
\end{figure}

\begin{figure}[ht]
    \centering
    \begin{subfigure}[b]{0.48\textwidth}
        \includegraphics[width=\textwidth]{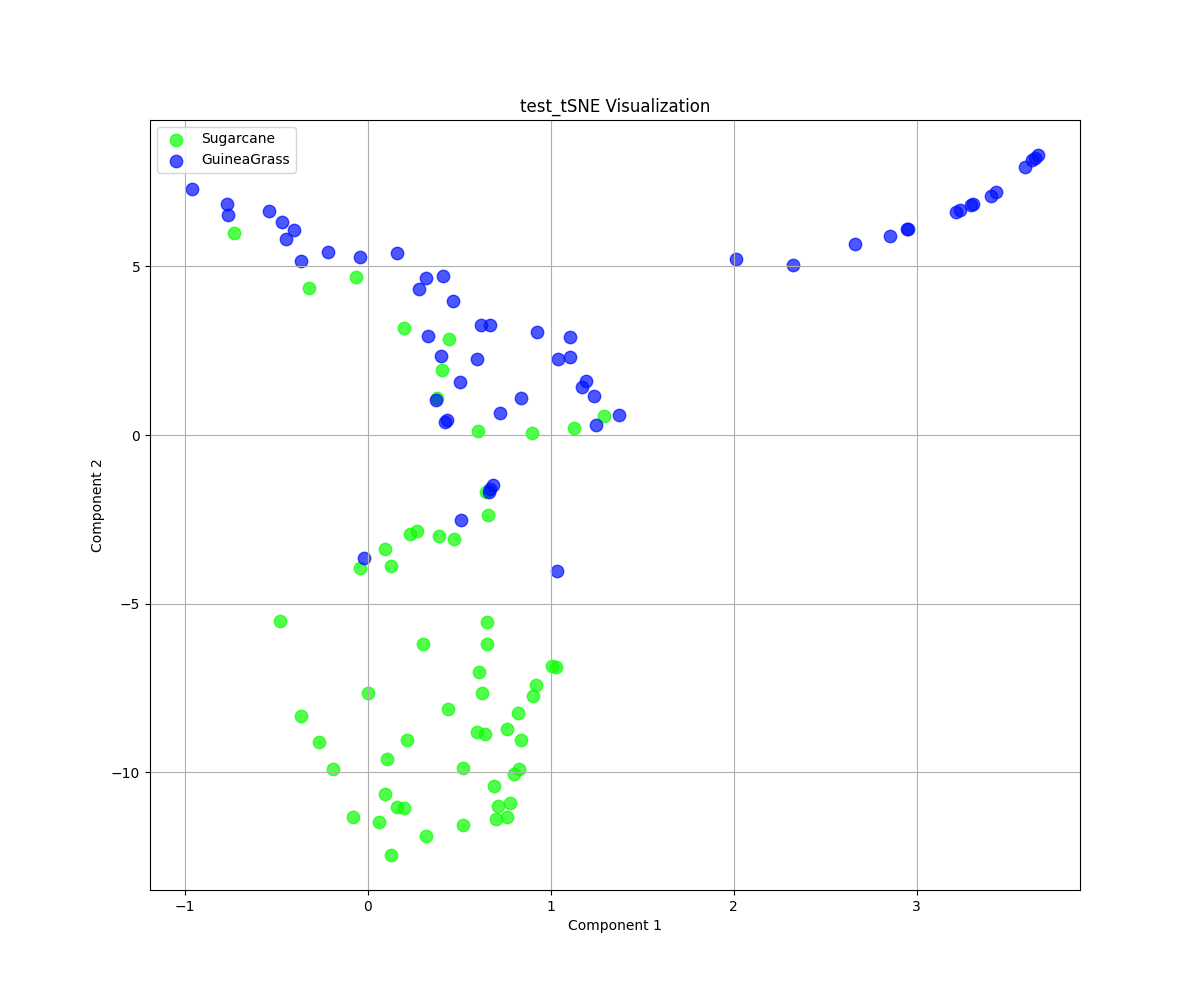}
        \caption{From-scratch training}
    \end{subfigure}
    \hfill
    \begin{subfigure}[b]{0.48\textwidth}
        \includegraphics[width=\textwidth]{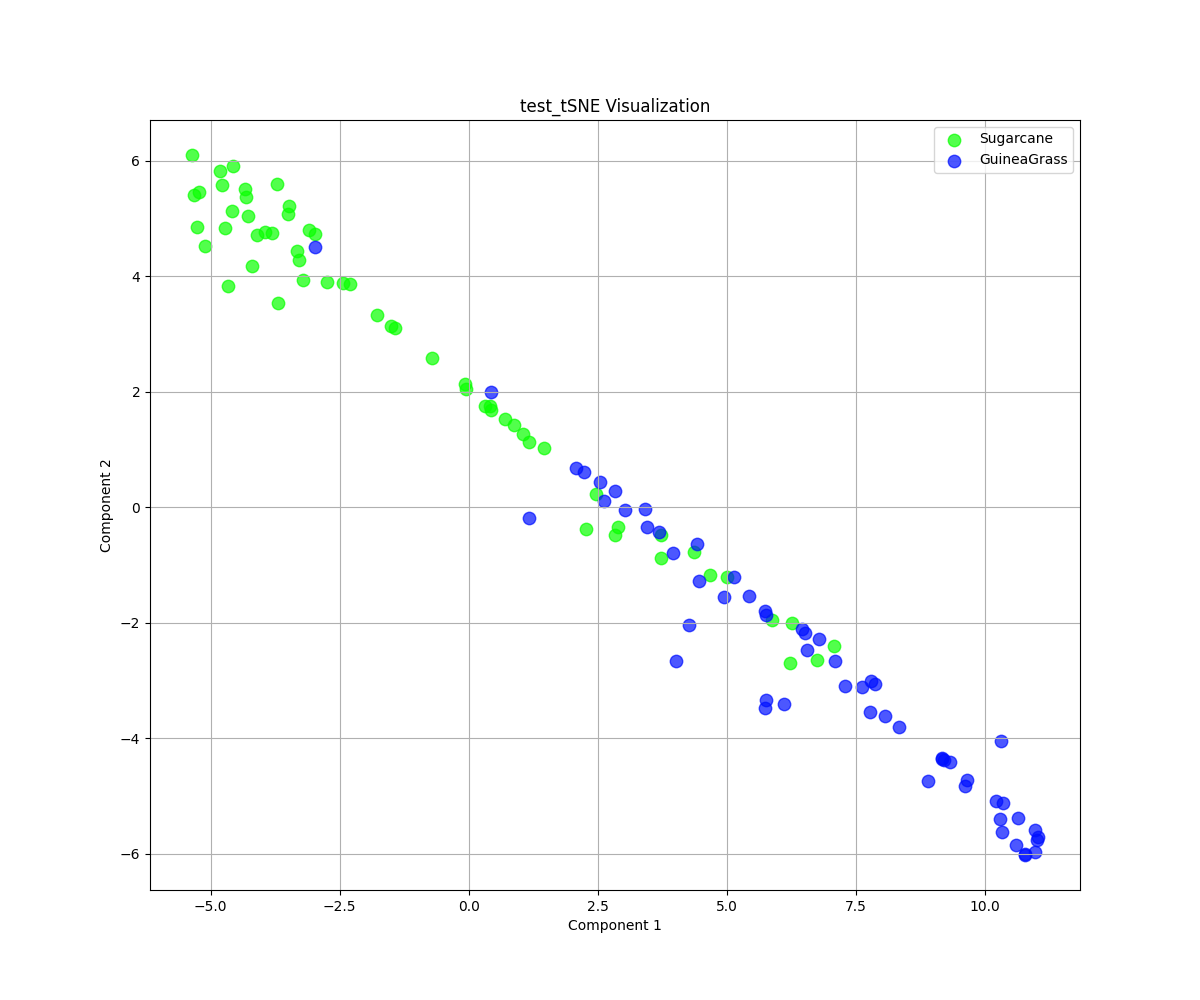}
        \caption{Pretrained initialization}
    \end{subfigure}
    \caption{t-SNE visualization of final-layer features. The pretrained model (b) shows slightly clearer cluster separation for sugarcane (green) and Guinea Grass (blue), aligning with its improved F1 score.}
    \label{fig:tsne}
\end{figure}

\paragraph{Analysis of Model Behavior.}
To better understand the model's decision-making process, we analyzed its performance using confusion matrices and t-SNE visualizations. The confusion matrices in Figure~\ref{fig:cmatrix} show that while both training strategies are effective, the pretrained initialization slightly reduces misclassifications. This finding is corroborated by the t-SNE projections of the final-layer embeddings (Figure~\ref{fig:tsne}), which reveal a slightly more distinct separation between the sugarcane and Guinea Grass clusters for the model initialized with pretrained weights.

% ----- Confusion Matrix and t-SNE Figures -----

\begin{figure}[ht!]
    \centering
    \includegraphics[width=0.98\textwidth]{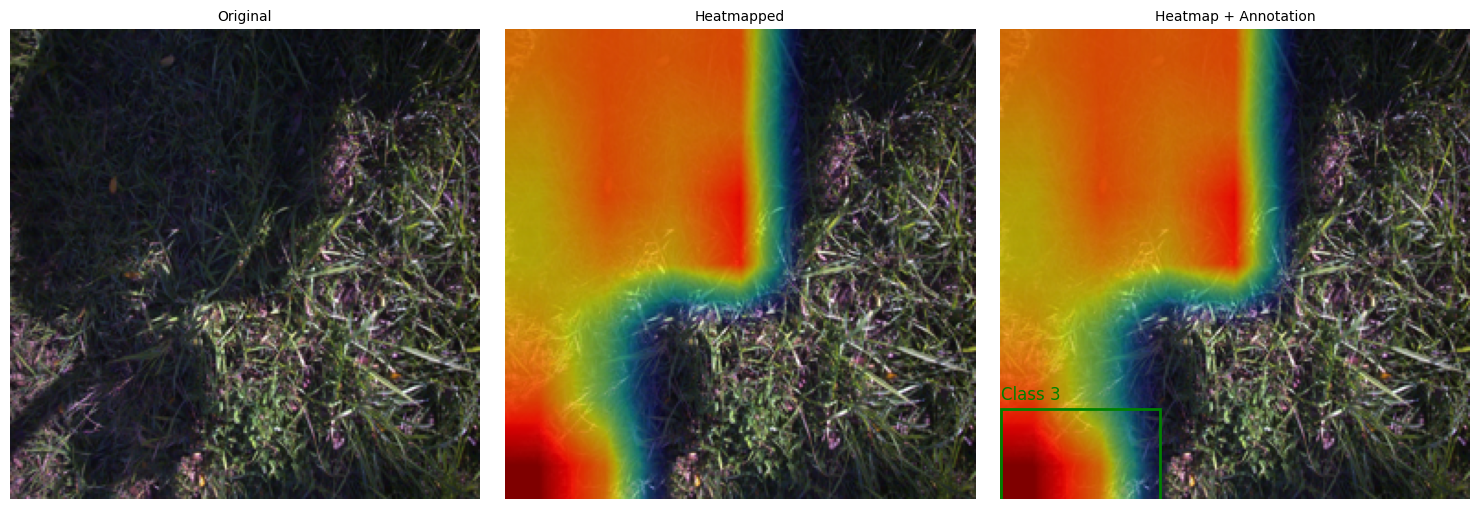}
    \caption{Grad-CAM visualization revealing the "shadow bias." The heatmaps show the model's attention (red areas) drifting towards background shadows instead of the target weed leaves, a key failure mode that guided our research towards object detection.}
    \label{fig:grad-cam}
\end{figure}

\paragraph{Diagnostic Finding: "Shadow Bias".}
Despite the high F1 score, we sought to understand the sources of remaining error. We employed Grad-CAM to visualize the classifier's attention, leading to a critical insight. As shown in Figure~\ref{fig:grad-cam}, the model frequently failed to focus on the actual morphological features of the weeds. Instead, its attention was often captured by high-contrast shadows or irrelevant background textures. This discovery of a "shadow bias" demonstrated that a robust solution required more than simple classification; it necessitated a model with stronger spatial understanding, motivating our subsequent shift to object detection.

% ----- Grad-CAM Figure -----

\subsection{Semi-Supervised Classification Results}
We extended our best fully supervised classification model to incorporate the \(\sim10,000\) unlabeled images via single-pass pseudo-labeling. A teacher model (SC2) was used to generate labels on the unlabeled set, with a high confidence threshold of 0.99 applied to ensure label quality. This process yielded 1,390 pseudo-labeled images. As shown in Table~\ref{tab:semisup-class-final}, this approach resulted in a marginal improvement in the test F1 score from 0.89 to \textbf{0.90}. While statistically positive, the modest nature of this gain further suggested that a classification framework might be reaching its performance ceiling for this complex task.

% ----- Semi-Supervised Classification Table -----
\begin{table}[ht]
\centering
\caption{Comparison of fully supervised vs. semi-supervised (SSC) classification performance. Adding pseudo-labeled data provides a marginal increase to the final test F1 score.}
\label{tab:semisup-class-final}
\begin{tabular}{lccc}
\toprule
\textbf{ID} & \textbf{Training Strategy} & \textbf{Val F1} & \textbf{Test F1} \\
\midrule
SC3 & Fully Supervised (Best) & 0.86 & 0.89 \\
SSC1 & Semi-Supervised (Student) & 0.85 & \textbf{0.90}\\
\bottomrule
\end{tabular}
\end{table}

\subsection{Object Detection Performance}
Motivated by the limitations of the classification approach, we transitioned to object detection to enable precise localization. All detection experiments were performed on full images, as our preliminary analysis showed that quadrant splitting degraded performance by truncating objects at boundaries.

\subsubsection{Fully Supervised Detection Baselines}
\begin{figure}[ht!]
  \centering
  \includegraphics[width=0.49\textwidth]{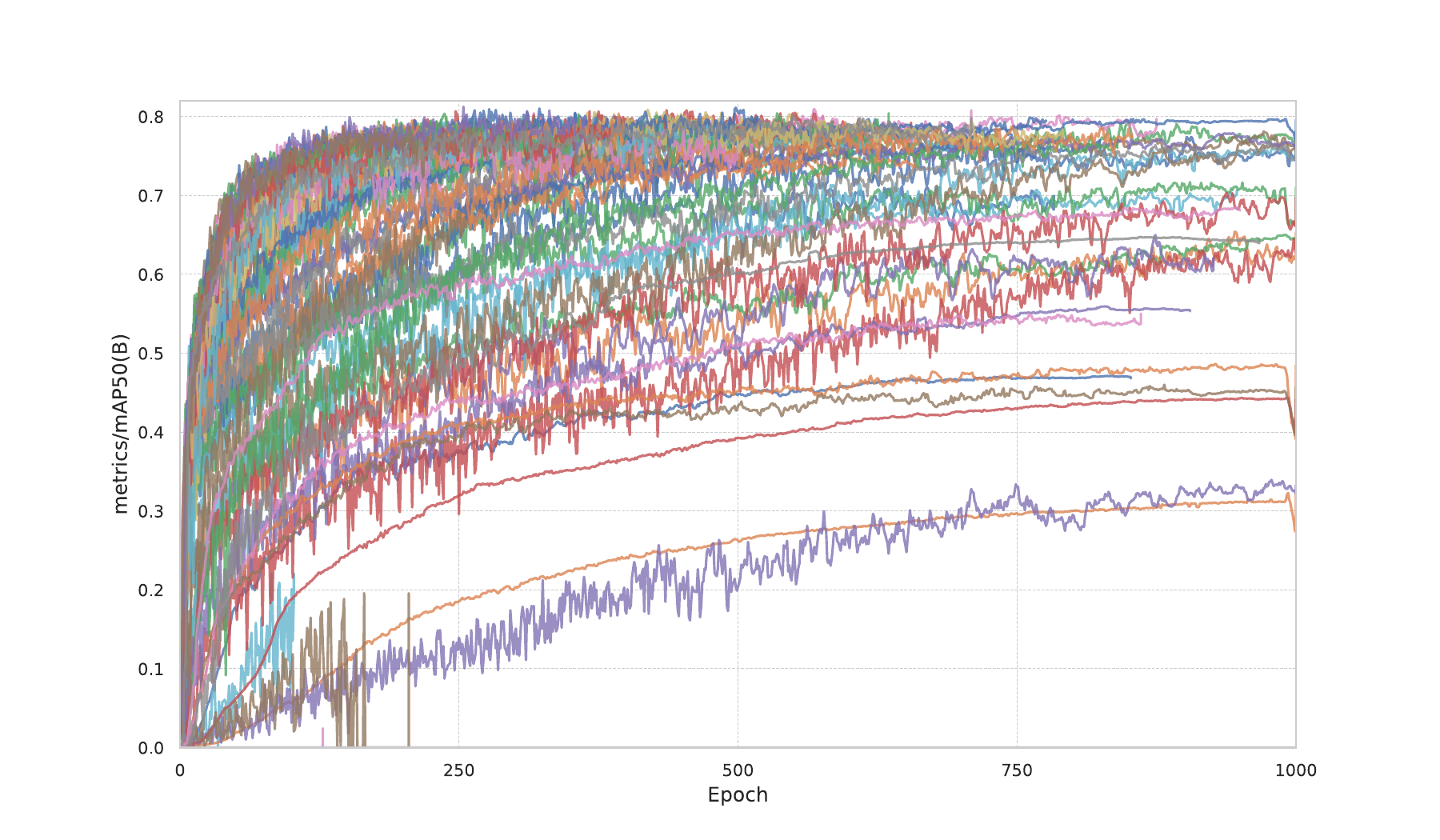}
  \hfill
  \includegraphics[width=0.49\textwidth]{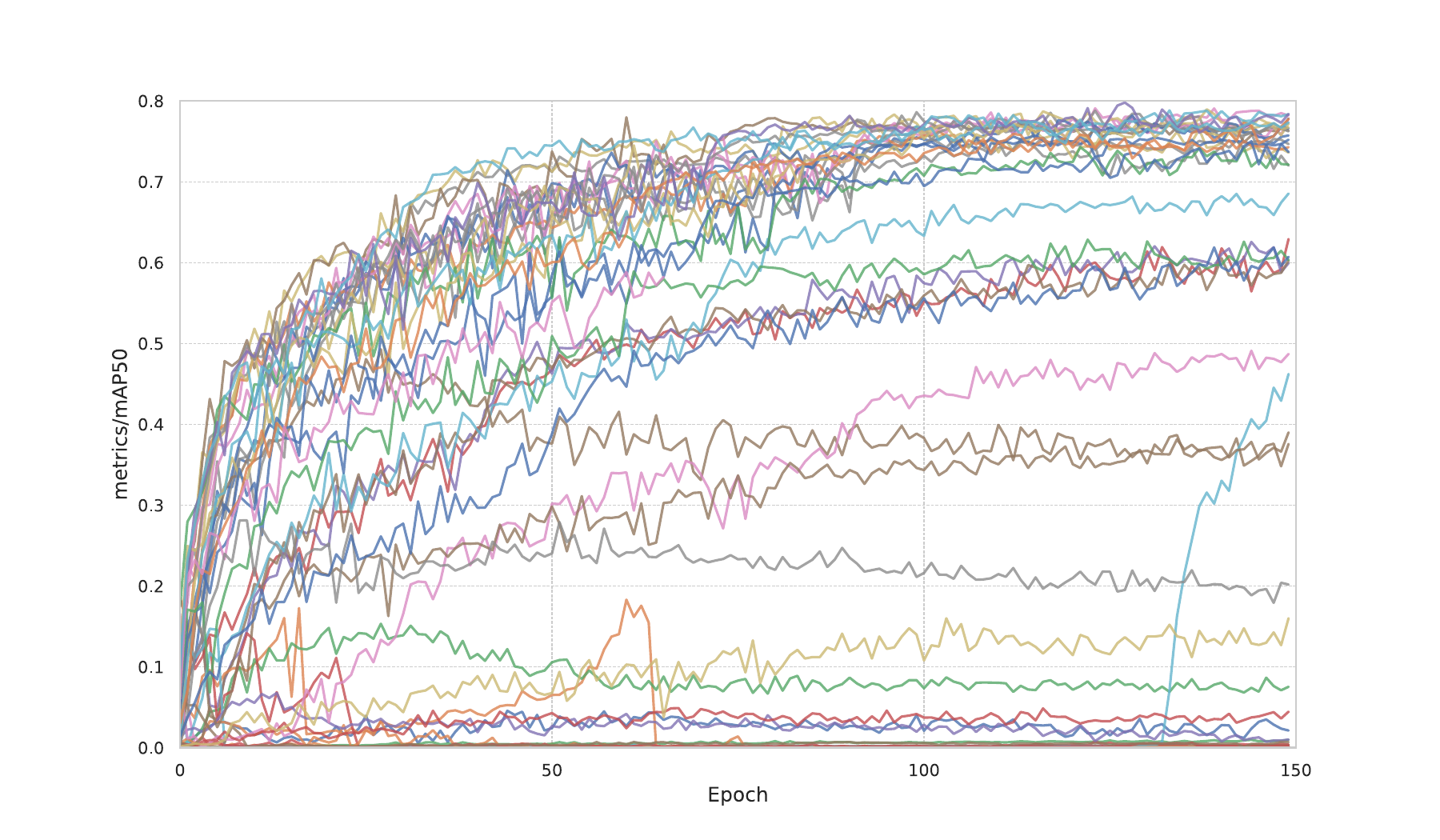}
  \caption{Validation mAP curves for YOLOv12-s (left) and RF-DETR (right) during supervised training. YOLO converges faster, while RF-DETR demonstrates a slower but steady improvement.}
  \label{fig:detect-training-curves}
\end{figure}

We first established supervised baselines for YOLOv12-s and RF-DETR on the combined A+B dataset. Training curves for representative runs are shown in Figure~\ref{fig:detect-training-curves}, illustrating that YOLOv12-s converges more rapidly than RF-DETR under our experimental conditions.
The comprehensive results are presented in Table~\ref{tab:det-ab-supervised}. The highly-tuned YOLOv12-s model (SD26) achieved a strong mAP@50 of 0.807 and an mAP@50-95 of \textbf{0.543}. The RF-DETR model (SD27), with minimal augmentation, achieved a respectable mAP@50 of 0.777, demonstrating the viability of Transformer-based approaches.

% ----- Detection Training Curves Figure -----

% ----- Supervised Detection Table -----
\begin{table}[ht!]
\centering
\caption{Fully supervised detection (SD) results on the combined Dataset A+B test set.}
\label{tab:det-ab-supervised}
\begin{tabular}{llcccc}
\toprule
\textbf{ID} & \textbf{Model} & \textbf{mAP@50} & \textbf{mAP@50-95} & \textbf{Precision} & \textbf{Recall} \\
\midrule
SD26 & YOLOv12-s & 0.807 & \textbf{0.543} & 0.804 & 0.771 \\
SD27 & RF-DETR   & 0.777 & 0.513 & 0.777 & 0.664 \\
\bottomrule
\end{tabular}
\end{table}

\subsubsection{Semi-Supervised Detection Enhancements}
We then applied our single-pass pseudo-labeling pipeline to the detection models. Based on our analysis of teacher model confidence distributions, we used a uniform 0.5 confidence threshold for YOLO, while for RF-DETR we employed a mixed-threshold strategy (0.8 for the difficult Guinea Grass class, 0.5 for others) to maximize label quality.

As shown in Table~\ref{tab:det-ab-semisup}, the impact of SSD was significant. The semi-supervised YOLOv12-s model (SSD8) improved its mAP@50 to {0.828}. More critically for practical applications, its {recall} saw a substantial increase from 0.771 to {0.782}. This reduction in false negatives (missed weeds) is a key outcome, as it directly translates to more effective field interventions. The semi-supervised RF-DETR model (SSD10) also showed improvement, reaching a mAP@50 of 0.783.

% ----- Semi-Supervised Detection Table -----
\begin{table}[ht!]
\centering
\caption{Semi-supervised detection (SSD) results. SSD significantly improves performance, most notably boosting the recall of the YOLOv12-s model.}
\label{tab:det-ab-semisup}
\begin{tabular}{llcccc}
\toprule
\textbf{ID} & \textbf{Model} & \textbf{mAP@50} & \textbf{mAP@50-95} & \textbf{Precision} & \textbf{Recall} \\
\midrule
SSD8  & YOLOv12-s & \textbf{0.828} & 0.529 & 0.814 & \textbf{0.782} \\
SSD10 & RF-DETR   & 0.785 & 0.507 & 0.785 & 0.675 \\
\bottomrule
\end{tabular}
\end{table}

\subsubsection{Qualitative Analysis and PubliSc Benchmark Validation}
\begin{figure*}[ht!]
  \centering
  \begin{subfigure}[b]{0.32\textwidth}
    \includegraphics[width=\textwidth]{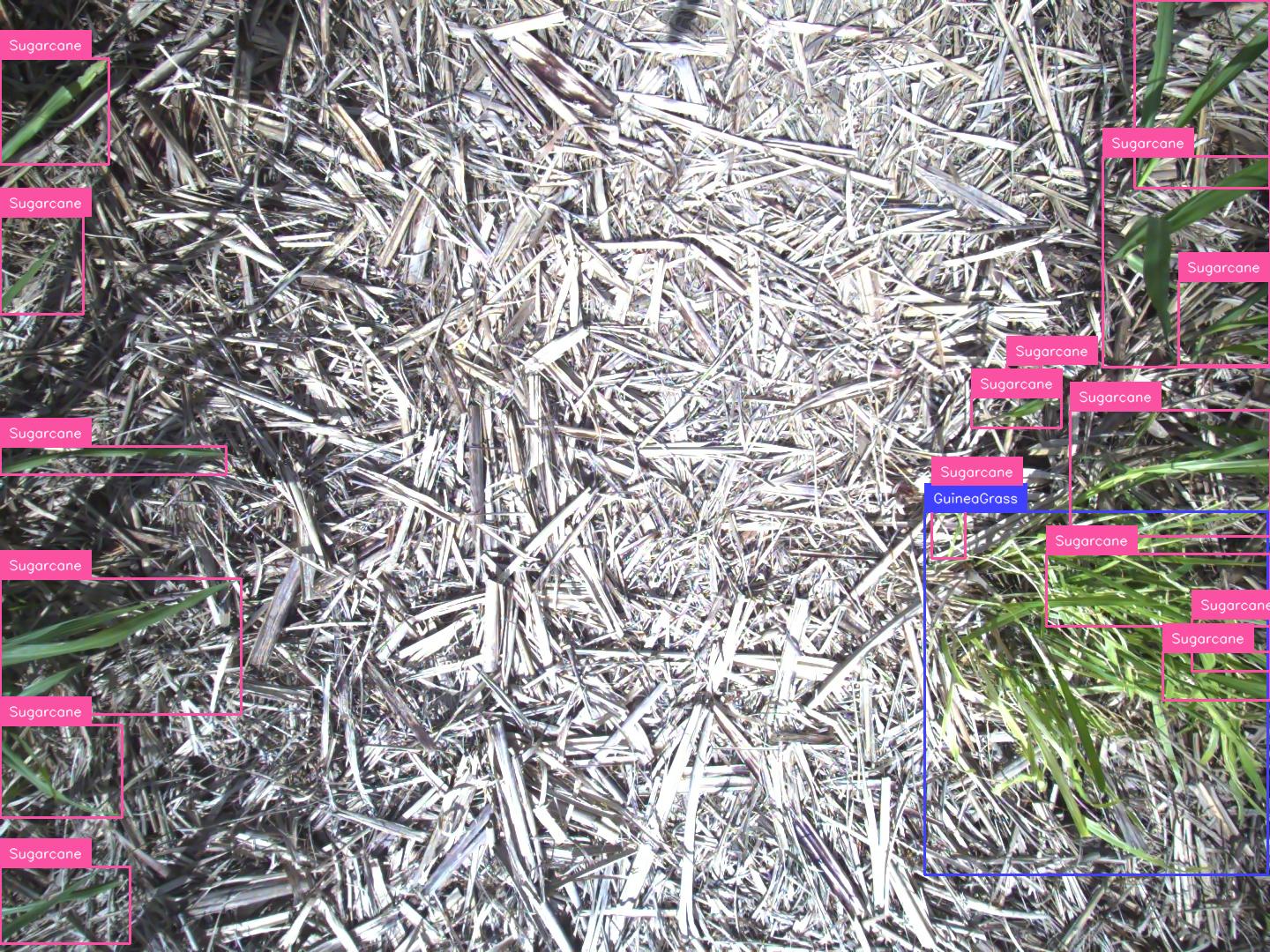}
    \caption{Ground Truth}
  \end{subfigure}
  \hfill
  \begin{subfigure}[b]{0.32\textwidth}
    \includegraphics[width=\textwidth]{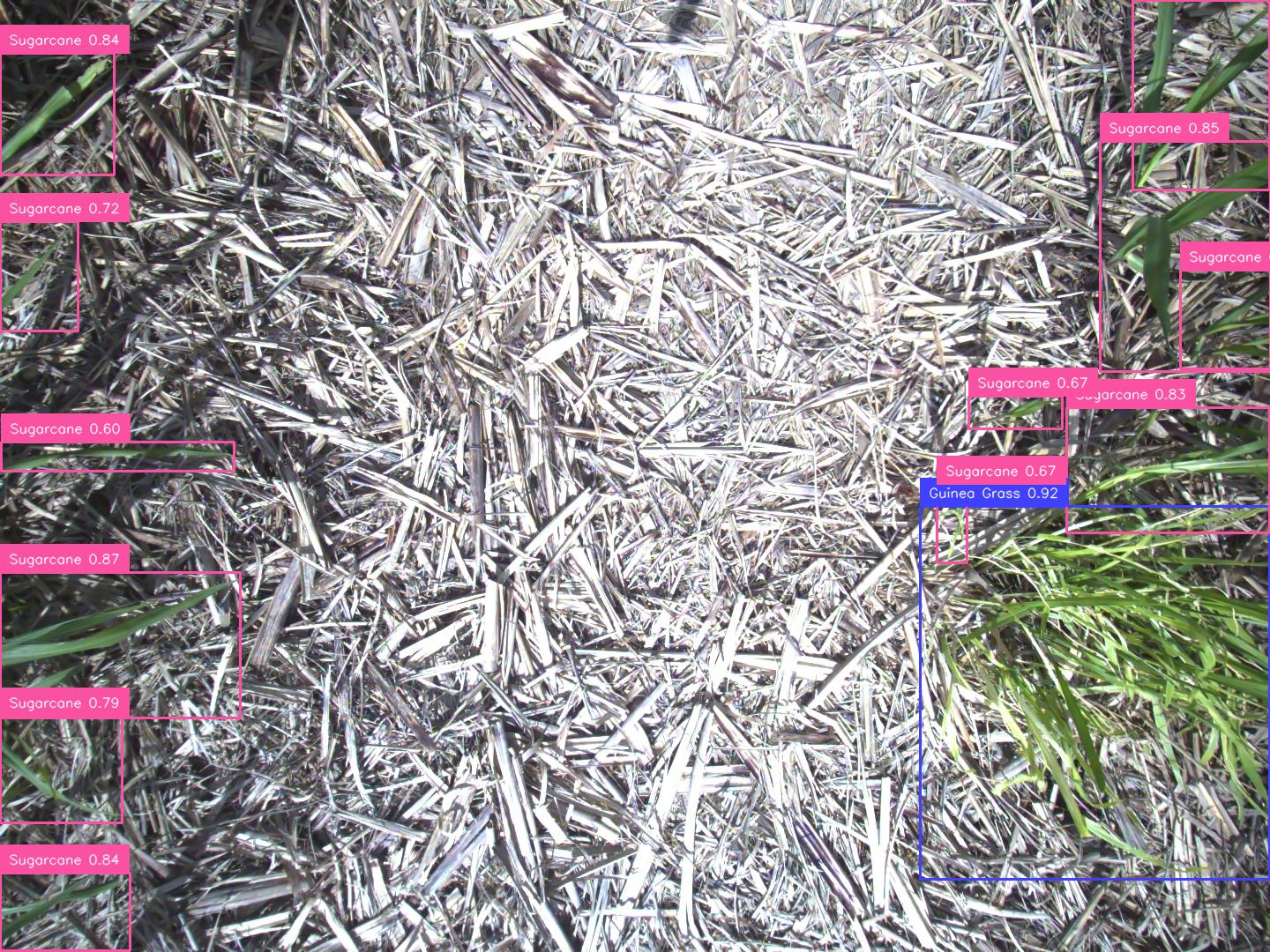}
    \caption{Semi-Supervised YOLO}
  \end{subfigure}
  \hfill
  \begin{subfigure}[b]{0.32\textwidth}
    \includegraphics[width=\textwidth]{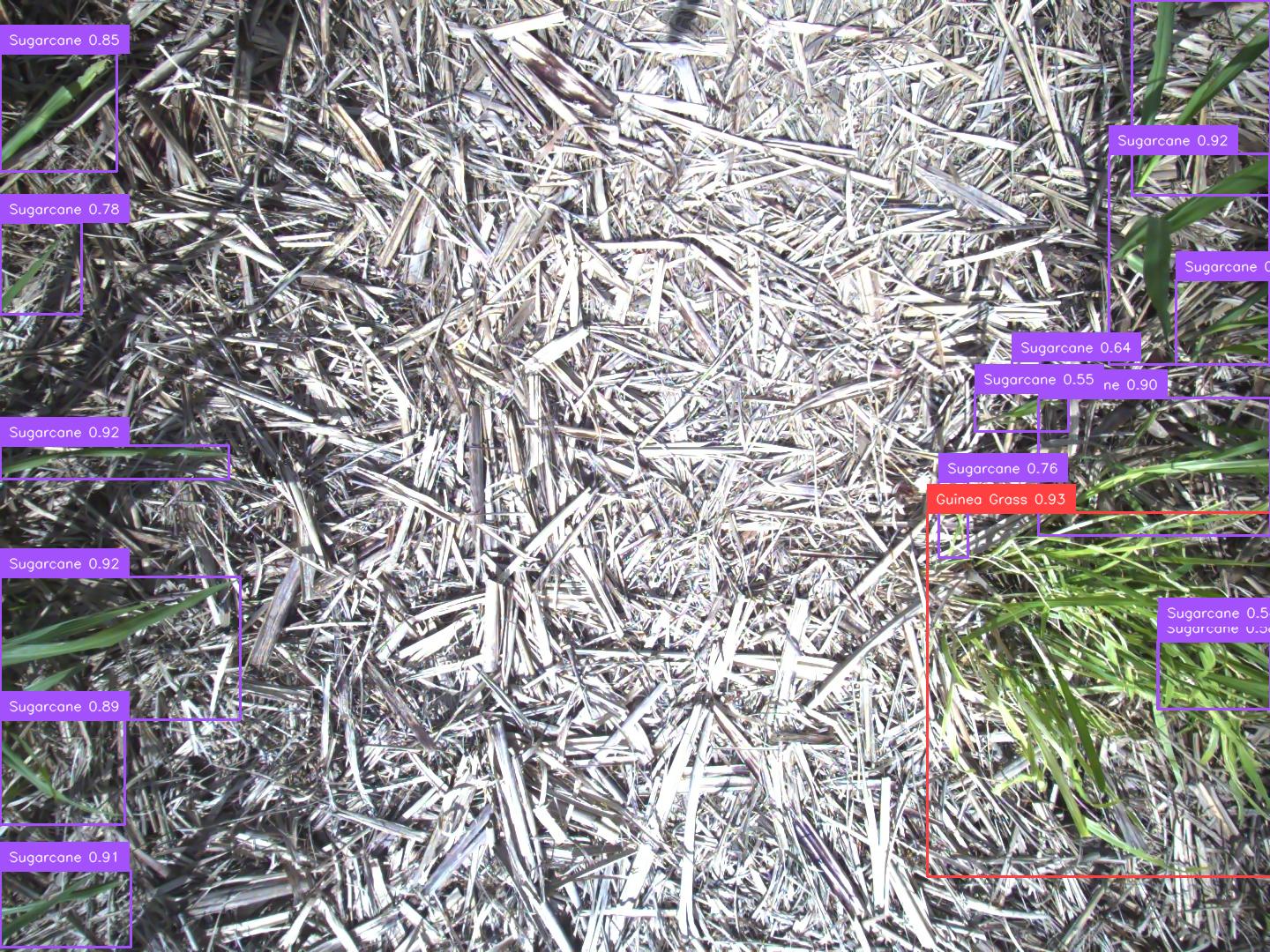}
    \caption{Supervised RF-DETR}
  \end{subfigure}
  \caption{Qualitative comparison on a challenging test image. (a) Ground truth. (b) The semi-supervised YOLO model correctly identifies all weed instances. (c) The supervised RF-DETR performs well but misses a partially occluded weed.}
  \label{fig:detect-qual-compare}
\end{figure*}

A qualitative comparison of the detection models is presented in Figure~\ref{fig:detect-qual-compare}. Finally, to validate the generalizability of our SSD methodology, we conducted an experiment on the public CropAndWeed dataset~\cite{Steininger2023WACV} under a low-data regime (10\% labeled data). As shown in Table~\ref{tab:public-results}, our SSD pipeline outperformed the supervised baseline, confirming its utility as a general technique for reducing annotation costs in agricultural vision.

% ----- Qualitative BBox and Public Benchmark Figures/Tables -----
% (Assuming fig:detect-qual-compare and tab:public-results are defined as in the previous response)

% (You can place the qualitative detection figure and the public benchmark table here, as they were in the previous response)

\begin{table}[ht]
\centering
\caption{Validation of the SSD pipeline on the public CropAndWeed dataset under a low-data regime. Our SSD approach significantly outperforms the supervised baseline, demonstrating its effectiveness in reducing labeling effort.}
\label{tab:public-results}
\begin{tabular}{lcc}
\toprule
\textbf{Training Method} & \textbf{Labeled Data Used} & \textbf{mAP@50} \\
\midrule
Supervised Baseline & 10\% & 0.90 \\ 
\textbf{Our SSD Pipeline} & \textbf{10\% (+90\% unlabeled)} & \textbf{0.91} \\ 
\bottomrule
\end{tabular}
\end{table}
%%%%%%%%%%%%%%%%%%%%%%%%%%%%%%%%%%%%%%%%%%%%%%%%%%%%%%%%%%%%

%%%%%%%%%%%%%%%%%%%%%%%%%%%%%%%%%%%%%%%%%%%%%%%%%%%%%%%%

\section{Discussion}\label{sec:discussion}

Our comprehensive investigation into weed detection reveals several key insights into the challenges and opportunities of applying deep learning in real-world agricultural settings. The results demonstrate a clear progression from initial, simpler models to more robust, data-efficient pipelines, with each stage providing valuable lessons.

\paragraph{From Classification to Detection: The Necessity of Spatial Awareness.}
Our initial approach using a quadrant-based ResNet-50 classifier achieved a high F1 score of 0.89. However, a purely metric-based assessment was misleading. Our critical finding, derived from Grad-CAM interpretability analysis (Figure~\ref{fig:grad-cam}), was the model's development of a "shadow bias." The model learned to associate high-contrast shadows with the presence of vegetation, a spurious correlation that limited its ability to generalize. This underscores a fundamental limitation of classification for dense-canopy tasks: it lacks the inherent spatial localization needed to distinguish target objects from their complex surroundings. This diagnostic step was crucial, as it justified our pivot to object detection, which is inherently better suited for precise localization and mitigating the impact of confounding background features.

\begin{figure}[t!]
    \centering
    \begin{subfigure}[b]{0.49\textwidth}
        \centering
        \includegraphics[width=\textwidth]{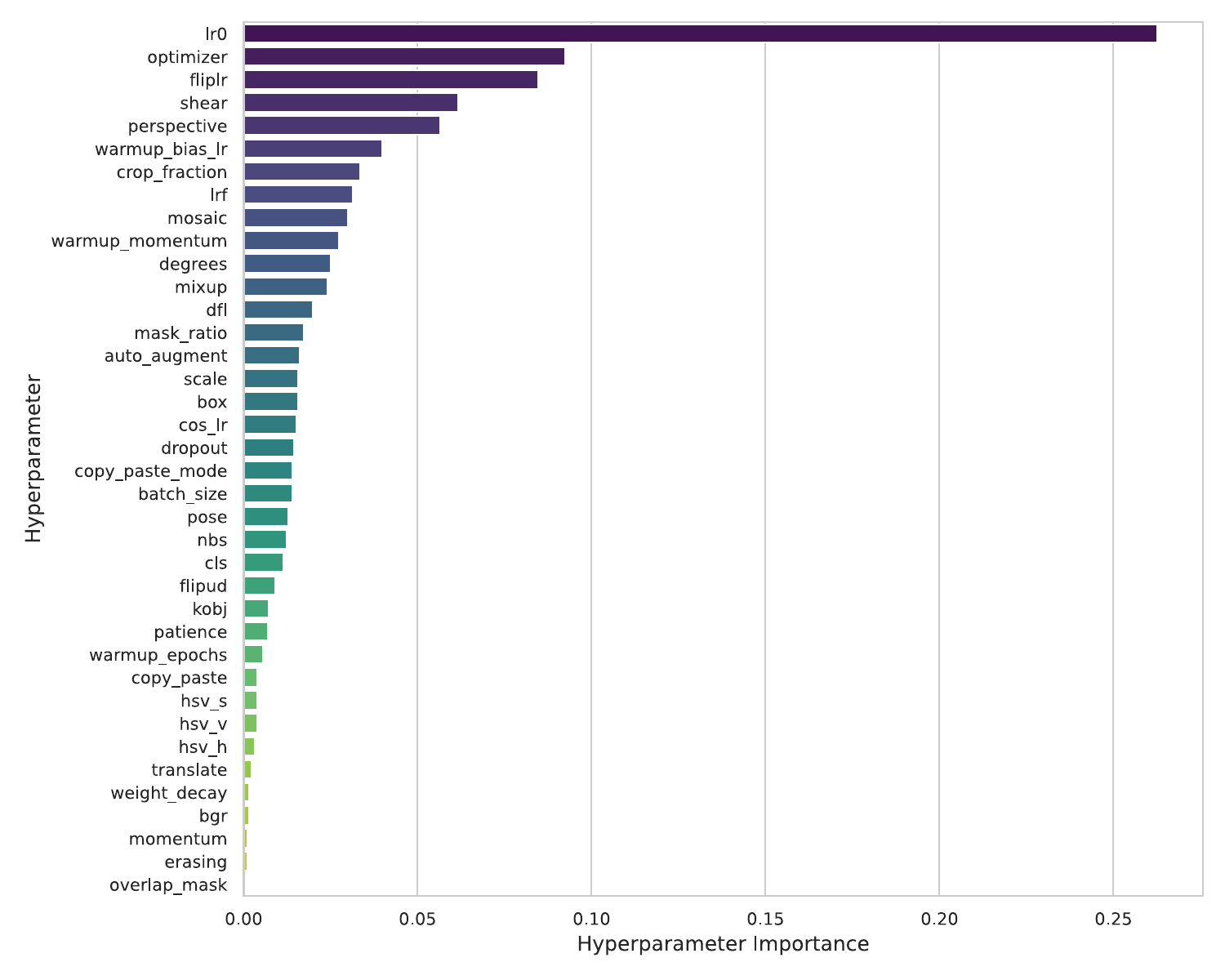}
        \caption{YOLO hyperparameter importance}
        \label{fig:yolo_sweep}
    \end{subfigure}
    \hfill
    \begin{subfigure}[b]{0.49\textwidth}
        \centering
        \includegraphics[width=\textwidth]{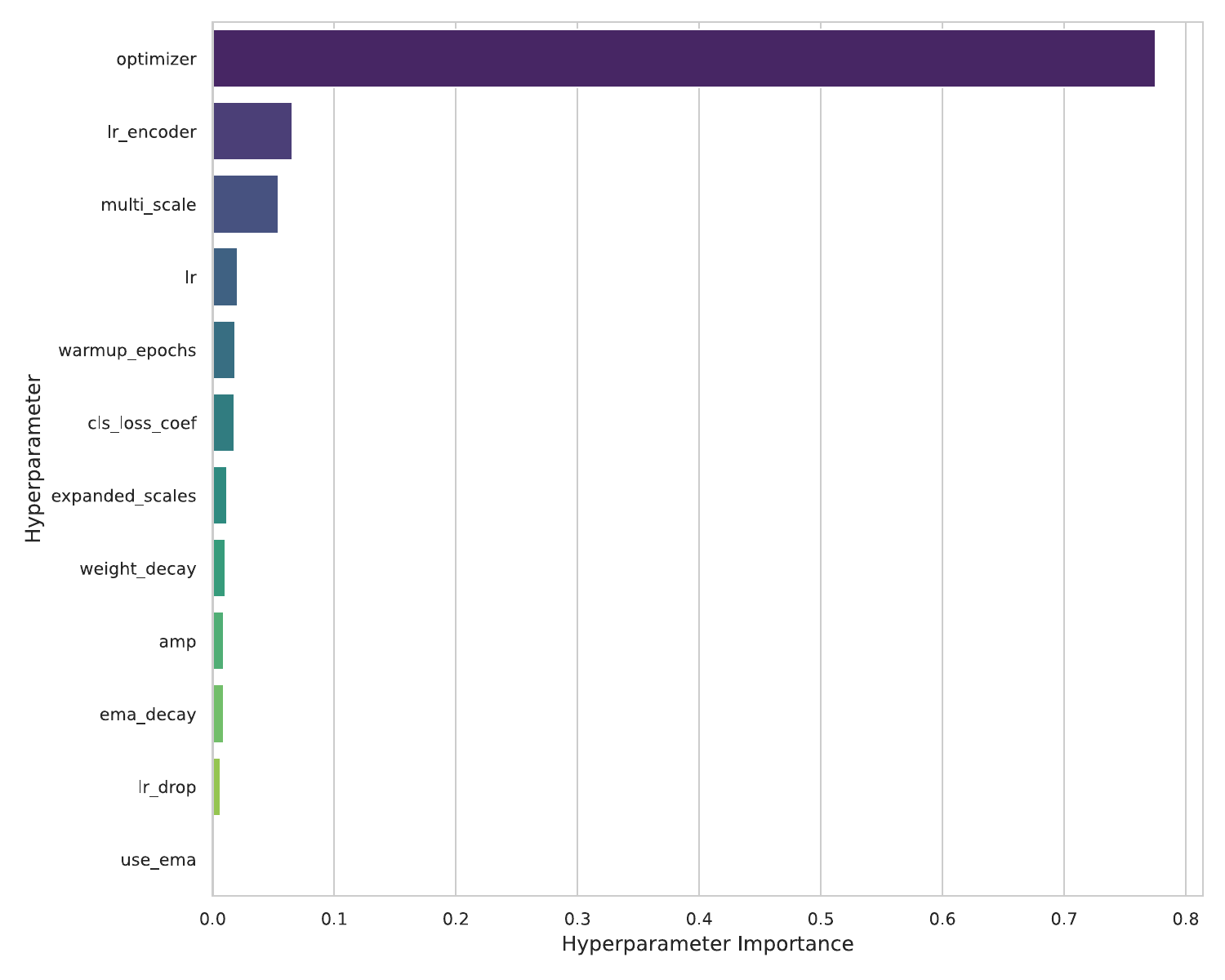}
        \caption{RF-DETR hyperparameter importance}
        \label{fig:detr_sweep}
    \end{subfigure}

\caption{
    \textbf{Optuna hyperparameter importance dashboards illustrating the differing sensitivities of YOLO and RF-DETR.}
    The length of each bar indicates the parameter's influence on the final mAP score during our extensive hyperparameter search.
    \textbf{(a)} For YOLOv12-s, the initial learning rate (\texttt{lr0}) is the most critical parameter, followed by the choice of \texttt{optimizer} and several geometric augmentations such as \texttt{fliplr} and \texttt{shear}. This highlights the model's high sensitivity to both the core training setup and the data augmentation pipeline.
    \textbf{(b)} In contrast, RF-DETR's performance is overwhelmingly dictated by the choice of \texttt{optimizer}, followed by architectural settings like the encoder learning rate (\texttt{lr\_encoder}) and multi-scale training.
    These findings confirm that while both models depend on fundamental training parameters, YOLO's performance is tightly coupled with its augmentation strategy, whereas RF-DETR's is more sensitive to its core architectural configuration.
}
    \label{fig:optuna-screen}
\end{figure}

\paragraph{Architectural Comparison: The Value of Specialization vs. Potential of Transformers.}
In our supervised detection experiments, the highly-tuned YOLOv12-s consistently outperformed RF-DETR. We attribute this performance gap not to an inherent superiority of CNNs, but to the maturity of the YOLO ecosystem. YOLOv12-s benefits from a powerful, built-in augmentation pipeline and has been subject to years of community-driven optimization for speed and accuracy. Our hyperparameter sweeps (Figure~\ref{fig:optuna-screen}) confirm this, showing that YOLO's performance is highly sensitive to geometric and color augmentations. In contrast, RF-DETR, a more recent architecture, required more careful tuning of its internal parameters (e.g., learning rates for the encoder) and was tested with minimal augmentation in our baseline. While our RF-DETR models achieved a respectable mAP of \(\sim0.78\), suggesting strong potential, closing the performance gap would likely require a more extensive, domain-specific augmentation strategy and hyperparameter search. This finding suggests that while Transformers are promising, their direct application to agriculture may require more tailored adaptation compared to well-established CNNs.

\paragraph{The Practical Impact of Semi-Supervised Learning.}
The most significant outcome of our study is the demonstrated value of semi-supervised learning. While the final mAP improvement from SSL was modest (from 0.807 to 0.828 for YOLO), the practical implication lies in the substantial boost in {recall}. In precision agriculture, particularly for automated spraying, a false negative (a missed weed) is far more costly than a false positive. A missed weed survives to compete with crops and produce seeds, leading to exponential future infestations. The \(\sim\)3\% absolute increase in recall for our YOLO model means significantly fewer weeds are missed in the field, a critical improvement for any real-world deployment. Furthermore, our validation on a public benchmark confirmed that our SSL pipeline is a generalizable technique that can drastically reduce annotation requirements, a key bottleneck in scaling agricultural~AI.

\paragraph{Limitations and Future Directions.}
Our study, while comprehensive, has several limitations that open avenues for future research.
\begin{itemize}
    \item {Proprietary Dataset:} While our dataset is uniquely challenging, it is not publicly available. Our validation on the CropAndWeed dataset helps mitigate this, but future work could involve creating and releasing a new, large-scale, high-density benchmark dataset for the community.
    \item {Simple SSL Approach:} We employed a single-pass pseudo-labeling strategy for its computational efficiency. More advanced SSL techniques, such as iterative teacher-student loops, consistency regularization, or domain adaptation methods, could potentially yield even greater performance gains and should be investigated.
    \item {Transformer Tuning:} As noted, our RF-DETR models were not subjected to the same degree of augmentation and tuning as our YOLO models. Future work should involve a more exhaustive hyperparameter search for Transformer architectures, including a wider range of augmentations, to conduct a more definitive comparison.
    \item {Edge Deployment:} While we focused on model accuracy, a full deployment pipeline must also consider inference speed and power consumption on edge devices. Future work should analyze these trade-offs, exploring model quantization and pruning to optimize our best-performing models for agricultural robots.
\end{itemize}

In summary, our work provides a clear pathway for developing robust weed detection systems, emphasizing the importance of diagnosing model failures and leveraging unlabeled data to overcome practical deployment barriers.
%%%%%%%%%%%%%%%%%%%%%%%%%%%%%%%%%%%%%%%%%%%%%%%%%%%%%%%%

\section{Conclusions}\label{conclusions}

In this work, we presented a comprehensive and systematic investigation into the detection of Guinea Grass in challenging, real-world sugarcane fields. Our study confronted the dual challenges of high crop-weed visual similarity and performance degradation due to environmental variability. By progressing from a simple classification pipeline to a more robust object detection framework, we demonstrated that a thorough, diagnostic-driven approach is critical for developing effective agricultural vision systems.

Our key contribution lies not in a single model, but in the methodology itself. We showed that interpretability tools like Grad-CAM are invaluable for uncovering subtle failure modes, such as the "shadow bias," which can mislead standard accuracy metrics. This insight guided our transition to object detection, where we benchmarked state-of-the-art CNN (YOLOv12-s) and Transformer (RF-DETR) architectures. While a highly-tuned YOLO model established the strongest supervised baseline, our most significant finding was the practical utility of semi-supervised learning. By leveraging a large pool of unlabeled data through a simple and efficient pseudo-labeling pipeline, we were able to enhance model performance, most notably achieving a crucial increase in recall. This directly translates to fewer missed weeds, a paramount objective for automated spraying systems.

Ultimately, our research underscores that the path to robust, field-deployable AI in agriculture requires more than just architectural improvements. It demands a holistic approach that includes rigorous diagnostics, the intelligent use of both labeled and unlabeled data, and a clear understanding of the specific failure modes relevant to the target environment. The framework and insights presented in this paper offer a practical roadmap for developing the next generation of scalable and effective weed management technologies, paving the way for more sustainable and efficient farming practices.
%%%%%%%%%%%%%%%%%%%%%%%
% Data Availability
%%%%%%%%%%%%%%%%%%%%%%%

\section*{Data Availability} \label{sec:Data_Availability}
The proprietary Guinea Grass dataset that supports the findings of this study was collected under a research agreement and is not publicly available due to commercial sensitivity and privacy restrictions. The public CropAndWeed dataset used for methodology validation is available at \cite{Steininger2023WACV}.
%%%%%%%%%%%%%%%%%%%%%%%%
\section*{Acknowledgement}
The authors gratefully acknowledge the funding support for this research, which was provided through the partnership between the Australian Government's Reef Trust and the Great Barrier Reef Foundation. We also extend our thanks to our project Industry partner, AutoWeed, specially, Dr Alex Olsen, for collecting the GG data used in this study and to James Cook University for providing the necessary infrastructure and computational resources to conduct this study.

%%%%%%%%%%%%%%%%%%%%%%%
% AUTHOR CONTRIBUTION
%%%%%%%%%%%%%%%%%%%%%%%
% \section*{Author Contributions}
% xxxxxx and reviewed the manuscript.

% \paragraph{Corresponding author}:\\Correspondence to alzayat.saleh@my.jcu.edu.au

%%%%%%%%%%%%%%%%%%%%%%%
% ADDITIONAL INFORMATION
%%%%%%%%%%%%%%%%%%%%%%%
\section*{Additional Information}
% To include, in this order: \textbf{Accession codes} (where applicable); \\
\textbf{Competing interests} The authors declare no competing interests.\\
All authors have read and agreed to the published version of the manuscript.
% \textbf{Ethical approval} This work was conducted with the approval of 
% The corresponding author is responsible for submitting a \href{http://www.nature.com/srep/policies/index.html#competing}{competing interests statement} on behalf of all authors of the paper. This statement must be included in the submitted article file.

%%%%%%%%%%%%%%%%%%%%%%%%
% \clearpage	
%  bibliography 
% \/ \/ \/ \/ 
% \ifCLASSOPTIONcaptionsoff
% \newpage
% \fi
% \IEEEtriggeratref{8}

% % references section
% \bibliographystyle{IEEEtran}
% % \bibliographystyle{IEEEtranN}
% \bibliography{references}
	
\printcredits

\bibliographystyle{unsrtnat}

\bibliography{references}

%%%%%%%%%%%%%%%%%%%%%%%%%%%%%%%%%%%%%%%%%%%%%%%%%%%%%%%%%%%%%%%%
\end{document}